\documentclass[journal]{IEEEtran}

\usepackage{graphicx}
\usepackage{amsmath}
\usepackage{hyperref}
\usepackage{color,soul}
\ifCLASSOPTIONcompsoc
  \usepackage[nocompress]{cite}
\else
  \usepackage{cite}
\fi
\ifCLASSINFOpdf
\else
\fi
\begin{document}
\title{Improving Solvability for Procedurally\\ Generated Challenges in Physical Solitaire\\Games Through Entangled Components}

\author{Mark~Goadrich
        and~James~Droscha%
\IEEEcompsocitemizethanks{\IEEEcompsocthanksitem M. Goadrich is with the Department
of Mathematics and Computer Science, Hendrix College, Conway,
AR, 72034.\protect\\
E-mail: see http://mark.goadrich.com/schedule.html
\IEEEcompsocthanksitem J. Droscha is with Glastyn Games.}%
}

\markboth{}%
{Shell \MakeLowercase{\textit{et al.}}: Bare Advanced Demo of IEEEtran.cls for IEEE Computer Society Journals}
\IEEEtitleabstractindextext{%
\begin{abstract}
    Challenges for physical solitaire puzzle games are typically designed in advance by humans and limited in number. Alternatively, some games incorporate rules for stochastic setup, where the human solver randomly sets up the game board before solving the challenge. These setup rules greatly increase the number of possible challenges, but can often generate unsolvable or uninteresting challenges. To better understand the compromises involved in minimizing undesirable challenges, we examine three games where component design choices can influence the stochastic nature of the resulting challenge generation algorithms. We evaluate the effect of these components and algorithms on challenge solvability and challenge engagement. We find that algorithms which control randomness through entangling components based on sub-elements of the puzzle mechanics can generate interesting challenges with a high probability of being solvable.
\end{abstract}

}

\maketitle

\IEEEdisplaynontitleabstractindextext
\IEEEpeerreviewmaketitle

\ifCLASSOPTIONcompsoc
\IEEEraisesectionheading{\section{Introduction}\label{sec:introduction}}
\else
\section{Introduction}
\label{sec:introduction}
\fi

\IEEEPARstart{P}{rocedural} content generation (PCG) algorithms are increasingly being used to derive new content for a wide variety of games\cite{PCGSURVEY}. Using the PCG taxonomy described by Togelius et al. \cite{SBPCG}, PCG can occur {\it offline} (beforehand) or {\it online} (dynamically during the game), the content can be {\it constructed}
by a system of rules, or use a {\it generate-and-test} process to winnow potential candidates for inclusion in the game, and the algorithm can be {\it deterministic} and fixed or {\it stochastic}, incorporating randomness. 

Following the puzzle terminology of Browne \cite{PUZZLENATURE}, such that there is a setter who creates challenges and a solver who solves them, we can apply the above PCG terminology to puzzle design. Setters typically {\it construct} their challenges using creative yet {\it deterministic} means. 
Researchers have explored using PCG to replace the setter, employing metaheuristics to find interesting challenges for deductive logic puzzles, ranging from Sudoku \cite{SUDOKU} to Nonograms \cite{NONOGRAM}. These algorithms construct their challenges offline or online, and guarantee they are solvable, but substitute {\it stochastic} algorithms for the creative human process. Khalifa and Fayek\cite{PUZZLELANG} investigated a combination of construction and generate-and-test PCG for Sokoban and related games within a genetic algorithm framework, and this approach was extended to Monte Carlo Tree Search by Kartal et al. \cite{SOKOBAN}.

A less-explored variety of puzzle with relation to PCG 
are physical solitaire games, for example
sliding block puzzles \cite{FIFTEEN} (including Rush Hour\footnote{https://www.thinkfun.com/products/rush-hour/}), 
and Hi-Q (generalized peg solitaire) \cite{PEG}. In these games, solvers must manipulate physical pieces to solve a challenge. Since the initial setup for these games must be executed by the solver, providing the solver with predefined challenges created offline and a solution book is common practice. PCG can also be applied to these games by, again, constructing challenges offline and guaranteeing they are solvable, as seen in recent work by Fogleman \cite{RUSHHOUR} and K{\"o}pp \cite{TANGRAM}. 

There are, however, alternative PCG approaches available for physical solitaire games, most popularly demonstrated by the card game Klondike Solitaire \cite{morehead2014complete}. In particular, this game uses an {\it online, stochastic, generate-and-test} PCG algorithm, which is as simple as shuffling the deck of cards at the start of the game. Also of note, the {\it test} portion of the generate-and-test algorithm is left to the solver as they play through the game. Wolter \cite{SOLITAIREVARIANTS} developed the Politaire system, and examines the effect of various shuffling algorithms across multiple solitaire card game variations. One variant called Thoughtful Solitaire, played such that all card locations are known to the solver at the beginning of the game, has been separately found to have between 82\% and 91.44\% of generated challenges solvable \cite{THOUGHTFUL}. 

While some claim that solitaire games with potentially unwinnable challenges are ``a rather sad form of amusing oneself," \cite{de1981pretzel} others find a ``catharsis [in] patience" even without perfect solvability \cite{morehead2014complete}. Thus, to address either side, when designing such a random setup process to assemble a challenge, two questions naturally arise for the designer:{\it What percentage of such challenges can be solved? Are the created challenges interesting?}\cite{MCPUZZLE} Also, since these puzzle challenges are instantiated by the solver, they are restricted to using only the components included in the puzzle. However, as much of PCG research involves digital games and puzzles, where algorithms are only limited by computational time and space constraints, there is little research into the relationship between the structural design of physical puzzle components and PCG. 

To understand this component/algorithm relationship, we examine three different styles of physical solitaire games with perfect information that incorporate {\it online, stochastic, generate-and-test} PCG algorithms for challenge setup: BoxOff \cite{BoxOffGAMES} , a token removal puzzle, Pretzel \cite{de1981pretzel} , a card arrangement puzzle, and Fujisan\footnote{http://www.ludism.org/ppwiki/Fuji-san}, a transportation puzzle.

Each of these puzzles includes a basic algorithm where simple components (tokens or cards) are randomly shuffled to create a challenge. We define these components to become {\it entangled} if their arrangement is no longer independent, either due to algorithmic constraints, or having been subsumed by new, larger components. We provide detailed descriptions of entangled game components and challenge setup algorithms for each game, then compare and contrast these using computational Monte Carlo simulations. Finally, we evaluate these algorithms with respect to solvability and interest. We find that entanglements that incorporate desirable constraints based on each puzzle's mechanics are more successful across both metrics.

\section{BoxOff}

\begin{figure}[t]
\centering
\includegraphics[width=6.8cm]{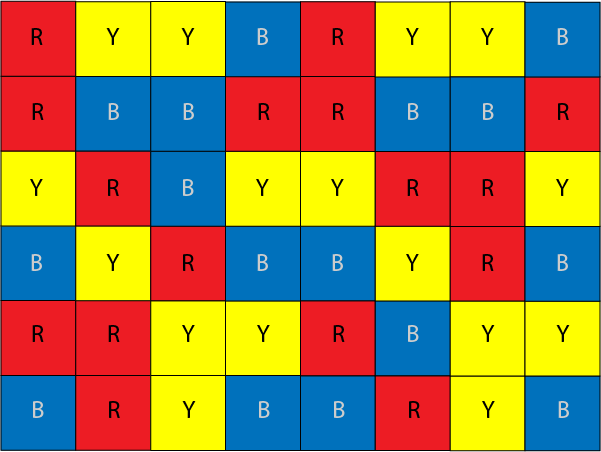}
\caption{A sample 6$\times$8 BoxOff challenge with three colors.}
\label{fig:boxoffbig}
\end{figure}
\noindent
BoxOff is a 2D token removal puzzle designed by Stephen Myers  \cite{BoxOffGAMES}. Challenges for this puzzle consist of a 6$\times$8 grid of squares filled with exactly 16 each of three colors, an example of which is shown in 
Figure \ref{fig:boxoffbig}. The challenge is solved when every square in the grid has been eliminated.  The solver can eliminate two squares of the same color at a time if either of the following two rules apply.
\begin{enumerate}
    \item The two squares are adjacent.
    \item All other squares inside the box that circumscribes the two squares have previously been eliminated.\footnote{Rule 1 is actually subsumed by this rule, however, we find it helpful to separate them into two cases to facilitate our component design discussions.}
\end{enumerate}

\begin{figure}[t]
\centering
\includegraphics[width=6.8cm]{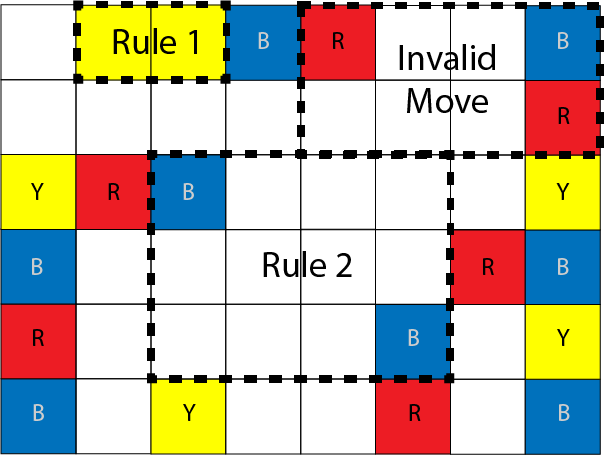}
\caption{Application of Rules 1 and 2 to an in-progress BoxOff challenge, and an illustrative invalid move.}
\label{fig:boxoffrules}
\end{figure}

Figure \ref{fig:boxoffrules} shows an in-progress BoxOff challenge. The solver has a few options available, two of which are highlighted in dashed boxes. They could eliminate the two yellow squares in the upper-left quadrant using Rule 1, or eliminate the two blue squares in the middle using Rule 2. An invalid move is also shown in the upper right quadrant, where Rule 2 would not apply because of the blue square in the circumscribed box around the two red squares.

Browne and Maire previously investigated altering the game design parameters for BoxOff using Monte Carlo simulation \cite{MCPUZZLE}. They explored multiple grid sizes and numbers of colors, $(h, w, c)$, where $h \times w$ is the size of the grid, and $c$ is the number of colors, ultimately finding that challenges generated for the original (6, 8, 3) puzzle configuration described by Myers were highly solvable and robust against random solvers. Other configurations, either of smaller board sizes, or more colors, exhibited poor solvability. In particular, we will examine the (4, 6, 4) puzzle, where challenge solvability was approximately 25\%, and the (6, 6, 6) puzzle, where challenge solvability was less than 5\%.

\subsection{Components and Algorithms}
\noindent
Here we discuss two different sets of physical components and their resulting PCG algorithms that can be used for challenge setup for Boxoff. 

The components of a physical BoxOff puzzle consist of circular tokens equal to the number of grid squares. These tokens are equally distributed among the number of colors. 
With these components, the solver can employ a simple shuffling algorithm to create a challenge.
\begin{quote}
    {\it \bf Shuffled:} Shuffle the color tokens randomly, and arrange them into a grid.
\end{quote}

We can use the multinomial theorem to determine that this method can create
$\frac{(hw)!}{c!^{k}}$, where $k=\frac{hw}{c}$. In the (4, 6, 4) puzzle this is $\approx 10^{11}$ possible challenges.  We acknowledge here and in subsequent calculations that rotational, horizontal, and vertical symmetry will alter the precise number of challenges, but not by an order of magnitude.

\begin{figure}[t]
\centering
\includegraphics[width=5.6cm]{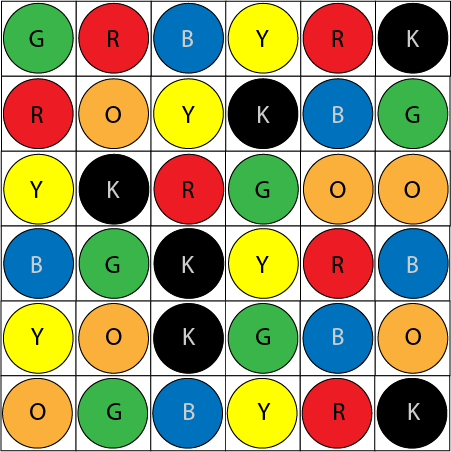}
\caption{A sample (6, 6, 6) BoxOff challenge, generated with a simple shuffle of the 36 color tokens. }
\label{fig:boxoff666}
\end{figure}

This shuffling works very well on moderate sized boards with few colors, achieving near 100\% solvability, but as demonstrated by Browne and Maire, it suffers when applied to more colors, as shown in Figure \ref{fig:boxoff666}. The uniform nature of the shuffling appears to prevent a critical mass of adjacent tokens of the same color. Such adjacent tokens are the only way Rule 1 above can be fulfilled so that more distant tokens can then be eliminated via Rule 2.

\begin{figure}[t]
\centering
\includegraphics[width=5.6cm]{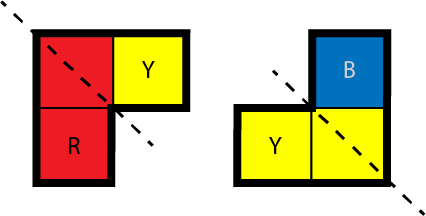}
\caption{Sample L-Tiles for BoxOff challenge creation. Tiles can be rotated 180 degrees, and the dashed line shows how they can be flipped. }
\label{fig:ltiles}
\end{figure}

To address this issue, we introduce an {\it entangled} abstraction for creating challenges using L-shaped tiles, as shown in Figure \ref{fig:ltiles}. These tiles allow us to entangle the colors on the tiles to create particular local color distributions.
Our goal in creating these tiles is to encourage more adjacent tokens of the same color. However, we also must avoid creating situations where all tiles are adjacent to tokens of the same color, else the puzzle becomes trivial to solve. 

\begin{figure}[t]
\centering
\includegraphics[width=8.8cm]{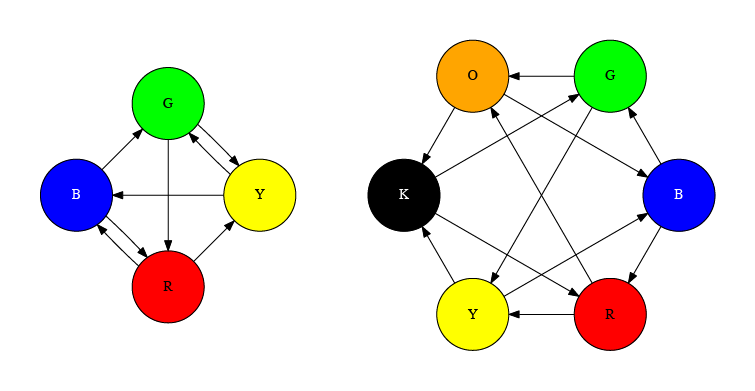}
\caption{Cycle of color connections used to construct L-Tiles for BoxOff challenges, (4, 6, 4) on the left, and (6, 6, 6) challenges on the right.}
\label{fig:boxoffcycle}
\end{figure}

For both the 4$\times$6 and 6$\times$6 grids, each tile is composed of exactly two squares of the same color, which we label $S$, and one square of a different color, labeled $D$. Each color is chosen as $S$ twice to ensure an equal distribution of the adjacent squares. We then distribute the third square colors using a cyclic arrangement of the colors, as shown in Figure \ref{fig:boxoffcycle}, where arrows are drawn from $S$ to $D$ for each tile in the set.  To increase the number of possible challenges, these tiles can be flipped to a mirror image along the center square of the L.

Each of these tiles has two possible orientations, for a total of $2^{t}t!$, where $t=\frac{hw}{3}$, for $\approx 10^{6}$ possible challenges in the (4, 6, 4) puzzle. 
The algorithm to create a challenge then is as follows:

\begin{quote}
    {\it \bf L-Tiles:} Shuffle and flip the tiles randomly, then arrange into a grid of connected 2$\times$3 subgrids. 
\end{quote}

\begin{figure}[t]
\centering
\includegraphics[width=5.6cm]{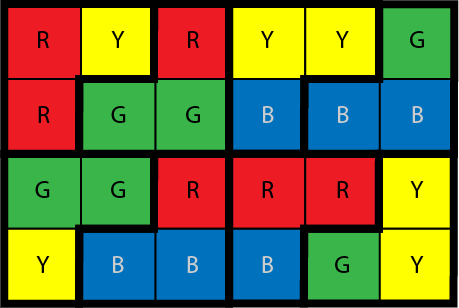}
\caption{A sample (4, 6, 4) BoxOff challenge, generated using L-Tiles. }
\label{fig:boxofftiles}
\end{figure}

Figure \ref{fig:boxofftiles} shows a (4, 6, 4) grid created with L-tiles. To physically play BoxOff with this tile setup, the solver will need to place neutral tokens on each square as they are eliminated, continuing until the board is full of tokens. We note that this provides an additional advantage of preserving the challenge throughout play; should the solver fail, all neutral tokens can be removed from the tiles and another attempt to solve the challenge can be made.

\subsection{Evaluation}
\noindent
We encoded a Monte Carlo BoxOff challenge generator using C\#, and implemented a breadth-first solver for BoxOff challenges. For each PCG algorithm, we generated 1000 random challenges. For each generated challenge in our trials, we recorded if the challenge was solvable. The code used for our simulations for this and subsequent puzzles is available on Github \footnote{http://github.com/mgoadric/entangled-components}. 
To enable testing for statistical significance on solvability between these algorithms, we divided our pool of generated challenges into 10 trials of 100 challenges. We repeated this for both the (4, 6, 4) puzzle and the (6, 6, 6) puzzle.
 
We will use two criteria to quantify each of the above components and algorithms.
First, we judge a PCG algorithm to be working well when a high percentage of generated challenges are solvable by our solver. Beyond solvability, we also wish for PCG algorithms to maintain or improve on the challenge interest for the basic \textbf{Shuffled} algorithm. 

\subsubsection{Solvability}

\begin{figure}[t]
\includegraphics[width=8.8cm]{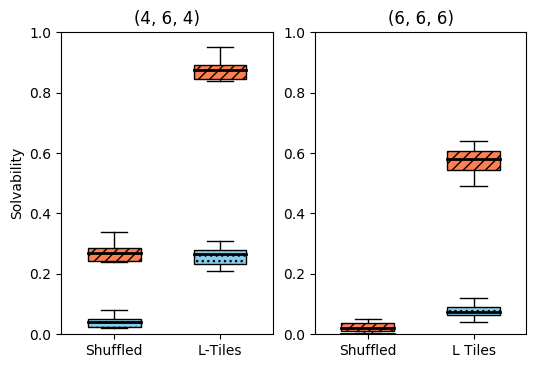}
\caption{Effect of challenge generation algorithms on solvability for BoxOff (4, 6, 4) and (6, 6, 6). $P(S_p)$ denoted in red with hatching, and $P(S_r)$ shown in blue with dots.}
\label{fig:boxoffsolvediff}
\end{figure}

Figure \ref{fig:boxoffsolvediff} shows the challenge solvability, in red with slashes, for the (4, 6, 4) BoxOff. This plot displays the distribution of the two PCG algorithms across the 10 trials in a box-and-whisker plot.  
We found the {\bf L-Tiles} setup method produces solvable challenges at a rate of 88\%, a marked increase over the {\bf Shuffled} probability of 27\%, with this difference being statistically significant.
Similarly, Figure \ref{fig:boxoffsolvediff} also shows the solvability distributions in red for the (6, 6, 6) puzzle. We again saw a drastic increase in solvability, moving from 3\% for {\bf Shuffled} up to 57\% for {\bf L-Tiles}. 

\subsubsection{Interest}

Following Browne and Maire \cite{MCPUZZLE}, if a challenge is solvable by the solver choosing random moves each time, then the challenge lacks interest. Denoting the probability of solvability using the breadth-first search solver as $P(S_p)$ and the probability of the random solver as $P(S_r)$, we can calculate $P(S_p) - P(S_r)$, the difference between the intelligent and random solvability. Figure \ref{fig:boxoffsolvediff} also shows, in blue with dots, the solvability rate for these algorithm when using a random solver. 

We see in both cases for {\bf L-Tiles}, there is an increase in the random solver performance over {\bf Shuffled}. However, there remains a significant gap between the general and random solvability in both puzzles. For the (4, 6, 4) puzzle, this difference is 62\%, and for the (6, 6, 6) puzzle, this difference is 50\%. We can conclude that while more easy challenges are generated from {\bf L-Tiles}, the large majority of generated solvable challenges involve interesting solutions.

\section{Pretzel}
\noindent
The Montana family of patience card games includes variations such as Gaps, Spaces, Vacancies, Clown, Paganini, Red Moon, and Blue Moon. One variant, which allows no re-shuffles or re-deals, is a puzzle described and named by de Bruijn \cite{de1981pretzel} called Pretzel solitaire.

To set up a Pretzel challenge, denoted as $(k, n)$, where $k$ represents the number of suits and $n$ represents the number of ranks in each suit, the solver shuffles a deck containing one card for each of $n$ ranks in $k$ suits. The shuffled cards are dealt face-up into a grid with $k$ rows, each with $n$ columns. The solver then removes the lowest ranked card of each suit (traditionally aces) and places them in a new column to the left of the existing grid in a prescribed order (traditionally for four suits, from top to bottom: spades[$\spadesuit$], hearts[$\heartsuit$], diamonds[$\diamondsuit$], clubs[$\clubsuit$]). Doing so vacates $k$ cells in the grid, thereafter called holes. 

The solver's goal is to arrange all cards into ascending sorted order, one suit per row, following a single rule:

\begin{enumerate}
    \item A card may be moved from any grid location into any hole only if the card being moved is of the same suit and exactly one rank higher than the card immediately to the hole's left.
\end{enumerate}

\begin{figure}[t]
\centering
\includegraphics[width=7cm]{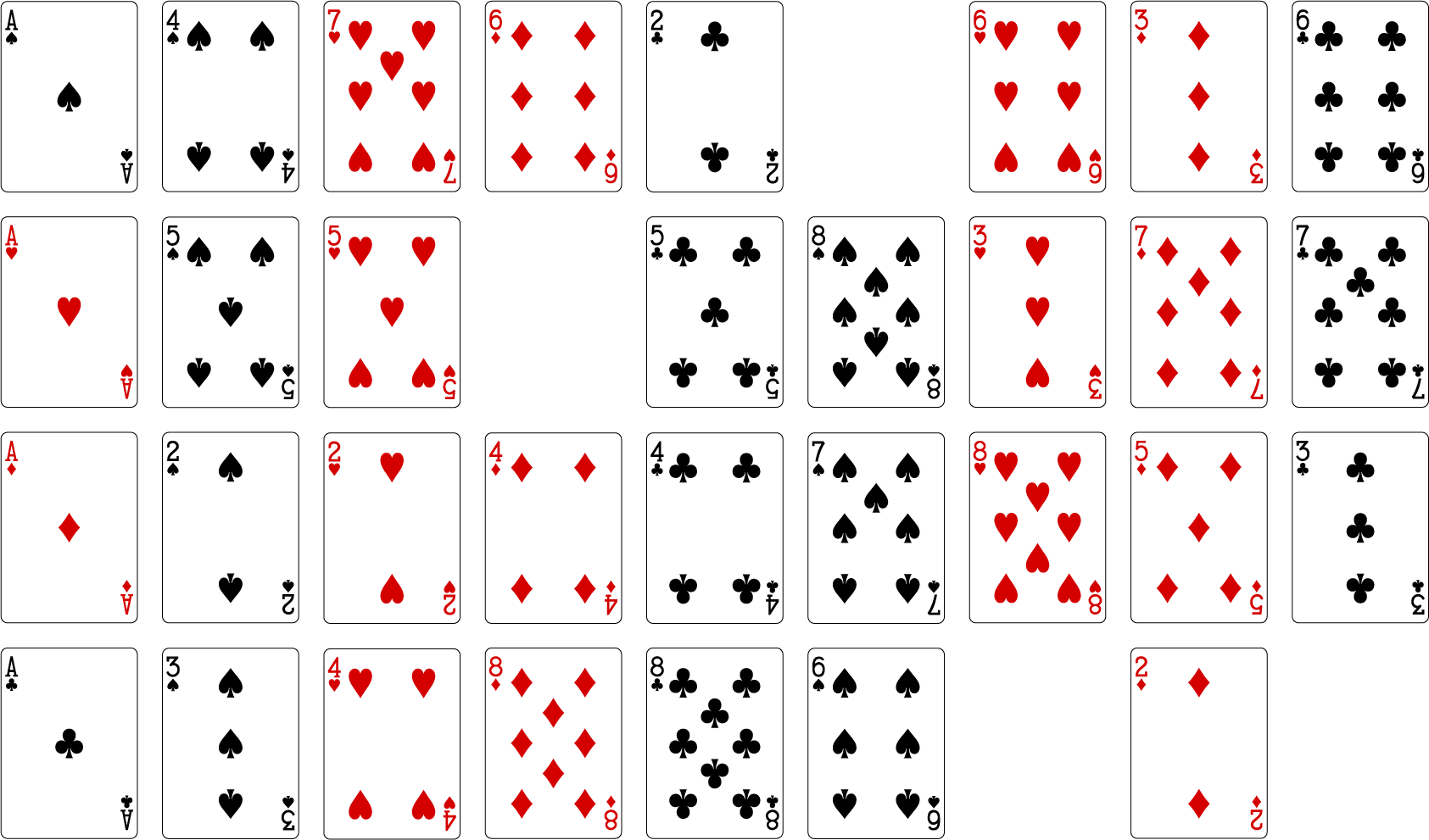}
\caption{Pretzel solitaire challenge with four suits and eight ranks, generated with the Sequential Suits method.}
\label{fig:pretzelsequential}
\end{figure}

\subsection{Components and Algorithms}
\noindent
Here we explore different PCG algorithms that can be used for Pretzel challenge setup. Each algorithm uses the features of the existing components, a standard French deck of cards, in different ways. We start with the traditional method of straightforwardly shuffling and dealing.

\begin{quote}
    {\bf Shuffled:} Shuffle the entire deck and deal all cards into the grid, left to right, top to bottom.
\end{quote}

The number of possible challenges that result from the standard shuffle is the total number of permutations of cards in the deck, namely $(kn)!$. For the (4, 4) case, this is $16! \approx 10^{13}$ distinct challenges.

In his paper, de Bruijn empirically established a solvability of 45\% for the (4, 4) case with a \textbf{Shuffled} setup. In an attempt to improve solvability with the next two setup algorithms, we introduce \textit{entanglement} of the cards by first dividing the deck into smaller decks according to suit. The two algorithms differ in how they distribute these \textit{entangled} components to complete the setup. Although the \textit{entanglement} is not preserved during play, the lasting effects of deck substitution during setup are nonetheless observable by measures similar to the other games we explored.

\begin{quote}
    {\bf Sequential Suits:} Divide the deck into $k$ smaller decks, each comprising all cards of a single suit. Shuffle each suit deck separately. Deal the top card from the first suit deck, then the top card from the second suit deck, and so on.  After the top card of each suit deck has been dealt, return to the first suit deck and deal the top card, then the top card of the second suit deck, and so on. Proceed in like fashion until all cards have been dealt, left to right, top to bottom.
\end{quote}

Figure \ref{fig:pretzelsequential} shows a sample Pretzel solitaire challenge with four suits and eight ranks, generated with the \textbf{Sequential Suits} method. The \textbf{Sequential Suits} setup algorithm generates $n!^{k}$ possible challenges. For the sake of comparison with \textbf{Shuffled}, the (4, 4) case with \textbf{Sequential Suits} provides $4!^{4} \approx 10^{5}$ distinct challenges.

Depending on the ratio of $k$ to $n$, challenges generated by this algorithm exhibit various distinct vertical or diagonal suit patterns across the play grid. After observing some gains in solvability arising from such patterns, we pushed the pattern to its logical limit:

\begin{quote}
    {\bf Banded Suits:} Divide the deck into $k$ smaller decks, each comprising all cards of a single suit. Shuffle each suit deck separately. Deal all cards from the first suit deck \textbf{top to bottom, left to right}. After each suit deck is exhausted, continue dealing with the next suit deck. Proceed in like fashion until all cards have been dealt, \textbf{top to bottom, left to right}.
\end{quote}

Dealing the cards from top to bottom, left to right is essential to the \textbf{Banded Suits} algorithm. In other words, the solver deals the first card into the upper-left location in the grid, then deals the next directly below the first in the same column, and so on. Only as each column is filled are cards dealt into the next column to the right. The result is $k$ bold, vertical suit bands.

As with \textbf{Sequential Suits}, distinct challenges created from \textbf{Banded Suits} is equal to $n!^{k}$. Also, note that \textbf{Sequential Suits} and \textbf{Banded Suits} produce identical suit patterns when $k = n$, but differing patterns for $k \neq n$.

\subsection{Evaluation}
\noindent
As with BoxOff, we encoded a Monte Carlo Pretzel challenge generator using C\#, and a breadth-first solver for Pretzel challenges. For each PCG algorithm, we generated 1000 random challenges. For each generated challenge in our trials, we recorded whether the challenge was solvable, and if so, we also recorded the minimum solution length found with our solver. To enable testing for statistical significance on solvability between these algorithms, we divided our pool of generated challenges into 10 trials of 100 challenges. 

\subsubsection{Solvability}

\begin{figure}[t]
\includegraphics[width=8.8cm]{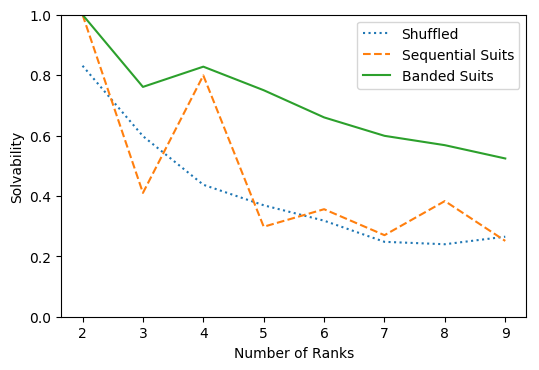}
\caption{Effect of challenge generation algorithms on solvability for Pretzel, four suits and 2-9 ranks.}
\label{fig:pretzelfullsolve}
\end{figure}

\begin{figure}[t]
\centering
\includegraphics[width=8cm]{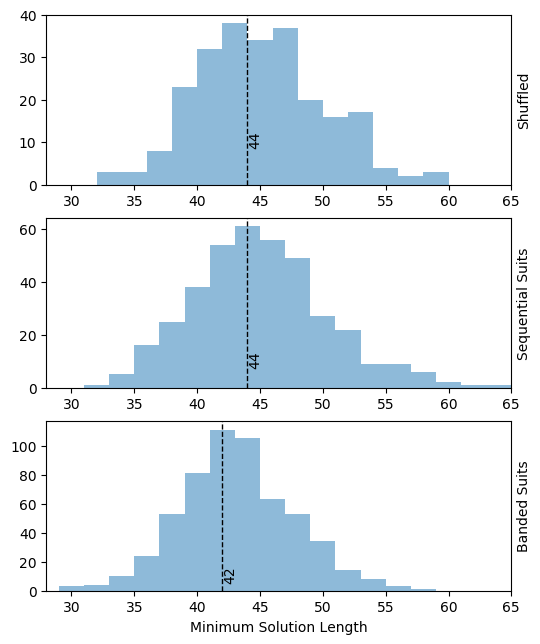}
\caption{Histograms showing the effect of Pretzel challenge generation algorithms on solution length, four suits and eight ranks.}
\label{fig:pretzellength}
\end{figure}

\begin{figure}[t]
\includegraphics[width=8.8cm]{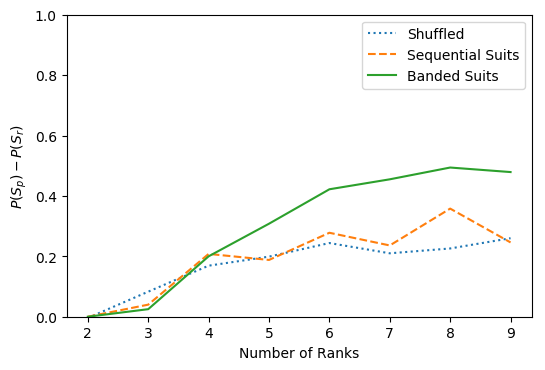}
\caption{Effect of challenge generation algorithms on $P(S_p) - P(S_r)$ for Pretzel, four suits and 2-9 ranks.}
\label{fig:pretzelinterest}
\end{figure}

Figure \ref{fig:pretzelfullsolve} compares the solvability of our three setup algorithms for four suits and two through nine ranks per suit. Solvability is displayed using the blue dotted line for \textbf{Shuffled}, the orange dashed line for \textbf{Sequential Suits}, and the solid green line for \textbf{Banded Suits}. Although the number of ranks is always a whole number, and intermediate, fractional rank values are therefore not sensible, we have used continuous lines in the figure to better depict overall trends for all algorithms tested and to highlight the oscillating nature of solvability for \textbf{Sequential Suits}.

We found that whether entanglement helped or hindered solvability when using the \textbf{Sequential Suits} setup algorithm is dependent on whether the number of ranks in the deck is even or odd. We also observed a marked increase in solvability with \textbf{Banded Suits} as compared to \textbf{Shuffled}, with the increase being statistically significant for all ranks tested. We include more details on the success of entangled components in Section \ref{entangled}.

The cost of solvability gains for \textbf{Banded Suits} is a decrease in minimum required solution length (in moves) which became statistically significant as the number of ranks increased. As a specific example, Figure \ref{fig:pretzellength} graphs the distribution of minimum required solution lengths recorded by our breadth-first solver for each of the setup algorithms in the (4, 8) case. The median minimum required solution lengths for \textbf{Shuffled} and \textbf{Sequential Suits} were identical at 44, while the median for \textbf{Banded Suits} dipped to 42. As with BoxOff, however, we judge the interest of challenges not primarily by their solution lengths, but by their resistance to random play.

\subsubsection{Interest}

Again denoting the probability of solving using the breadth-first search solver as $P(S_p)$ and the probability of the random solver as $P(S_r)$, we calculated $P(S_p) - P(S_r)$ using each of the setup algorithms across two through nine ranks in four suits. The results are shown in Figure \ref{fig:pretzelinterest}. The blue dotted line displays this value for \textbf{Shuffled}, the orange dashed line represents \textbf{Sequential Suits}, and the solid green line shows \textbf{Banded Suits}.

Here we see that \textbf{Banded Suits} exhibits the desired resistance to random play. For (4, 4) Pretzels, \textbf{Sequential Suits} and \textbf{Banded Suits} perform similarly. This is expected since they produce identical suit patterns for this case. But at five ranks and beyond, intelligent play outpaces random play by a significant margin as compared with \textbf{Shuffled} and \textbf{Sequential Suits}. We conclude that despite a slight decrease in minimum required solution length for \textbf{Banded Suits}, interest is retained.

\section{Fujisan}
\noindent
Fujisan was created specifically for the piecepack game system \cite{GAMESYSTEM}. In Fujisan, a solver must find a way to cooperatively move four Shinto Priests to the top of Mt. Fuji through incremental steps up the mountainside. Functionally, the area of play consists of a grid of spaces arranged into two rows by twelve columns. Each space contains a single value in the range of 0 to 5, inclusive. The two middle columns together comprise the mountain {\it summit}, while each other column forms a {\it step} of the mountain. Four pawns, representing the Priests, start off the mountain, just outside the two columns furthest from the summit.

\begin{figure}[t]
\centering
\includegraphics[width=7.8cm]{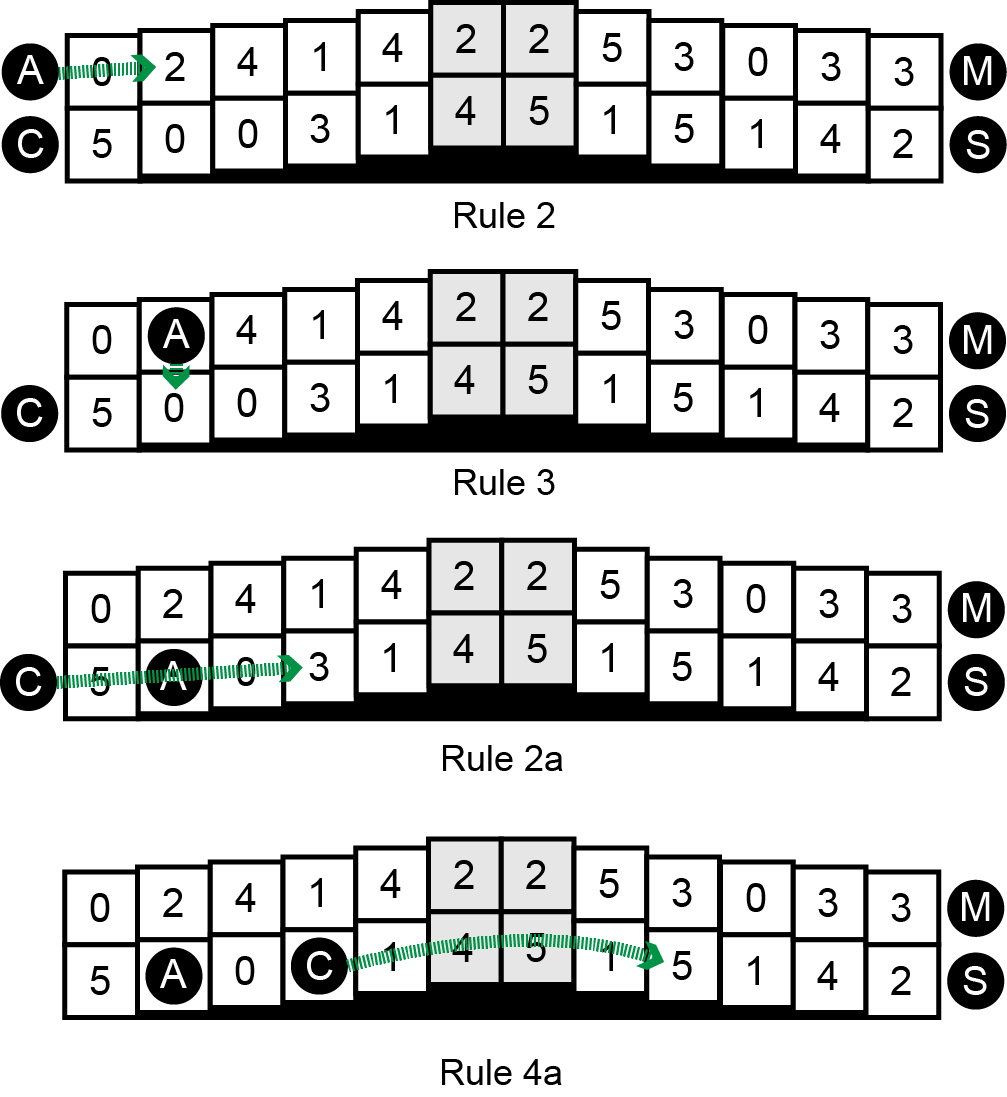}
\caption{The start of a solution demonstrating the rules of Priest movement in Fujisan, with move notation followed by the matching rule. }
\label{fig:priestrules}
\end{figure}

The goal of the solver is to move Priests one at a time until all four are at the summit. A Priest can be moved according to the following rules:  

\begin{enumerate}
\item No more than one Priest may occupy a space at any given time.

\item A Priest may move onto a space if that space's value matches the number of unoccupied spaces the Priest must move in a straight line, left or right, to get there (including the destination space itself, but not including the Priest's starting space). %
\begin{enumerate}
\item Occupied spaces (containing intervening Priests) are not counted when determining the distance from a Priest to a given space. %
\end{enumerate}
\item A Priest may move freely up and down between the two spaces of any given step of the mountain. %
\begin{enumerate}
\item A Priest's first move from the starting position must land on the mountain; that is, the Priest cannot move up or down while on the ground.
\end{enumerate}

\item A Priest that lands on the mountain's summit can no longer move left or right, but may still move freely up or down within the column.\footnote{The original rules of Fujisan also allowed a Priest on the summit to freely move left and right at the summit. The summit rule as written here was made for the Engraved Tiles version, and we used this formulation of the rule for the computational simulations of all versions discussed.}
\begin{enumerate}
\item A Priest may pass over the summit as part of a move.
\end{enumerate}
\end{enumerate}

Figure \ref{fig:priestrules} shows a visual example of how these rules can be used to begin solving a sample challenge. We denote the goal summit spaces in gray.

\subsection{Components and Algorithms}
\label{section:pcgalgs}
\noindent
Here we explore three different sets of physical components and their resulting PCG algorithms that can be used for challenge setup for Fujisan. 

The piecepack is a set of board game parts that can be used to design and play a wide variety of games\cite{GAMESYSTEM}. The mountain in Fujisan was constructed with game tiles each marked with a 2$\times$2 grid. The values were added using 24 round coins, which represented the cross product of two sets: suits (sun, moon, crown, arms) and values (0, 1, 2, 3, 4, 5). Values are indicated on the front of the coin, while suits are found on the back.

We examine first a simple algorithm that can make use of the piecepack coin components to generate randomness. 

\begin{quote}
    
  {\it \bf Shuffled:} Shuffle the 24 coins face-down. For each space on the board, randomly select one coin and place it face-up on this space.
  
\end{quote}

We can use the multinomial theorem to determine that this method can create
$\frac{24!}{4!^{6}} \approx 10^{14}$ possible challenges.  We acknowledge here and in subsequent calculations that rotational, horizontal, and vertical symmetry will alter the precise number of challenges, but not by an order of magnitude.

However, if two 0 coins end up placed in the same column, then it becomes impossible to move a Priest onto that column. This creates holes in our challenges and reduces the number of solvable setups. 
More importantly, when both spaces of either of the summit columns contain 0s, the challenge becomes impossible to solve.

The original published Fujisan ruleset was devised to address the issue of double 0 steps, adding the constraint that each step must have two different values. This was achieved by using the suit information on the backs of the coins to \textit{entangle} the coins as they are placed, similar to the \textbf{Sequential Suits} method for Pretzel.

\begin{quote}
    
  {\it \bf Piecepack:} Shuffle the 24 coins face-down, and separate into four groups based on their suit. Then repeatedly place two coins on the two right-most available spaces, choosing from each of the suits in turn (sun, moon, crown, arms).
\end{quote}

With each space limited to choosing from a particular suit, the \textbf{Piecepack} algorithm will generate $6!^4 \approx 10^{10}$ possible challenges. This algorithm will guarantee there are no double numbers on a step, thus eliminating the double 0 issue noted above. 

\begin{figure}[t]
\centering
\includegraphics[width=3cm]{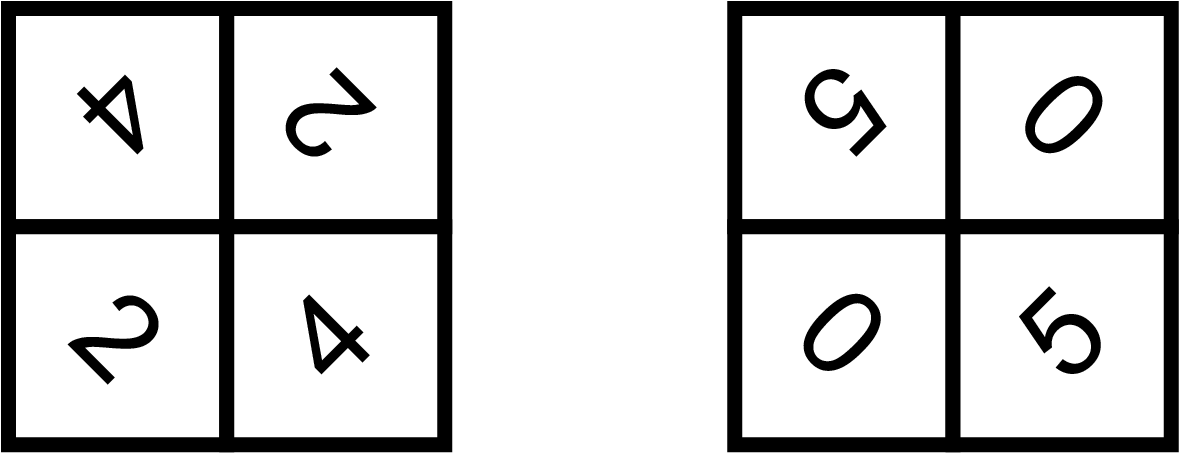}
\caption{Sample Engraved Fujisan Tiles.}
\label{fig:engravedsample}
\end{figure}

\begin{figure}[b]
\includegraphics[width=8.8cm]{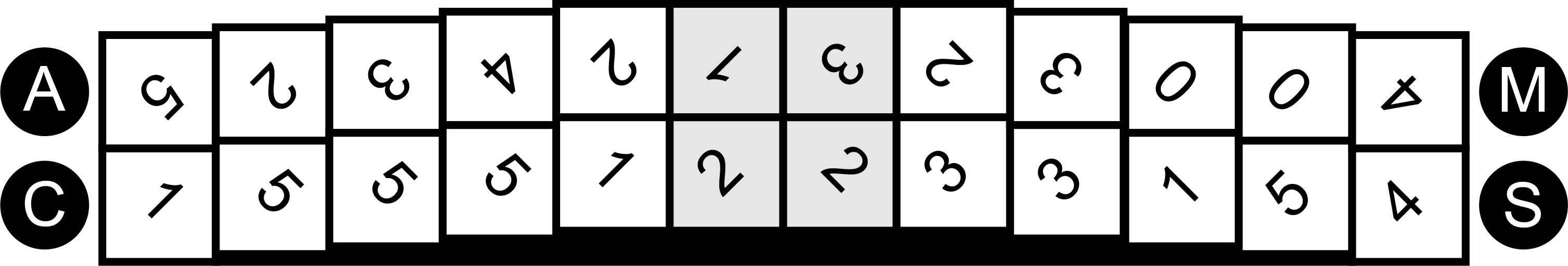}
\caption{A sample Fujisan challenge from the Engraved Tiles algorithm.}
\label{fig:tileexample}
\end{figure}

Another way to entangle the components in Fujisan is to combine the values with the 2$\times$2 tiles, engraving numerals onto the spaces, similar to the \textbf{L-Tiles} method for BoxOff. Here, we explore creating tiles with every possible pairing of values 0 through 5, including pairing a value with itself, and repeating these values diagonally on the tiles. Example tiles of this style are shown in Figure \ref{fig:engravedsample}. We remove the 0:0 pairing, since it can create unsolvable challenges, leaving 20 tiles.

\begin{quote}
    
  {\it \bf Engraved Tiles:} Shuffle the tiles face-down. Then, assemble the mountain by turning tiles face-up, using six for the bottom layer, five for the next layer, then four, then three, and finally two. The summit will be the center four spaces.
\end{quote}

This further constrains each pair of numbers to appear no more than once in the puzzle, except for the top two tiles. There are 20 possible tiles, and only 10 of them can be seen once the puzzle is constructed, as shown in Figure \ref{fig:tileexample}. 15 of these tiles have two possible orientations, for a total of $\sum_{i = 5}^{10}{15 \choose i}\binom{5}{10 - i}2^{i}10! \approx 10^{13}
$ possible challenges.

Furthermore, a standard double-six domino set, which includes 28 dominoes, 
can be used as entangled components. If we eliminate those dominoes that include a 6, along with all doubles, we are left with 15 dominoes. 

\begin{quote}
    
  {\it \bf Dominoes:} Shuffle the dominoes face-down. Place 12 of these dominoes face-up in a row to create the mountain. Place a face-down domino on each side of the mountain to denote the starting locations for the Priests. Place the remaining face-down domino horizontally in the middle to raise up the two central dominoes, denoting the summit.
\end{quote}

\begin{figure}[b]
\includegraphics[width=8.8cm]{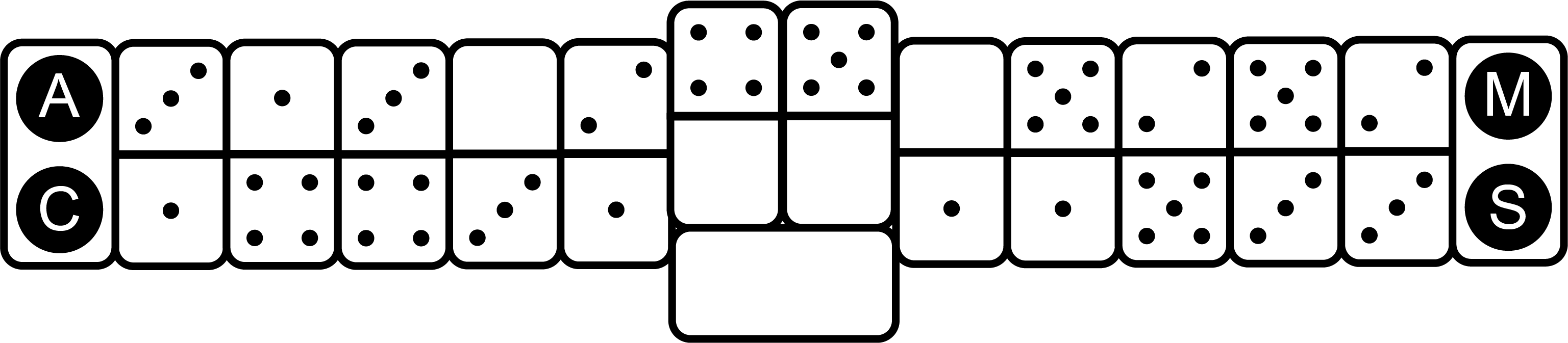}
\caption{A sample Fujisan challenge from the Dominoes algorithm.}
\label{fig:dominoexample}
\end{figure}

This constraint is similar to \textbf{Engraved Tiles}, but with a subset of the value pairs, thus a different probability on their selection.  Additionally, unlike \textbf{Engraved Tiles}, the summit values are distinct from the two steps closest the summit.
With 15 possible dominoes, only 12 of them are used in the challenge, as shown in Figure \ref{fig:dominoexample}. Each of these dominoes has two possible orientations, for a total of ${15 \choose 12}2^{12}12! \approx 10^{14}$ possible challenges. 

\subsection{Evaluation}

 \noindent

\begin{figure}[t]
\includegraphics[width=8.8cm]{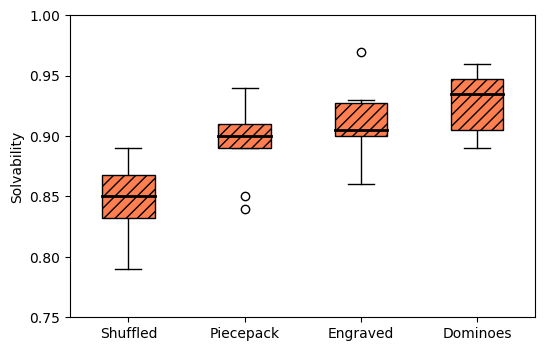}
\caption{Effect of Fujisan challenge generation algorithms on solvability.}
\label{fig:strategycomp}
\end{figure}

We encoded a Monte Carlo Fujisan challenge generator using C\#, along with an A* solver for Fujisan challenges. Our admissible heuristic for A* is the number of empty spaces on the summit. For each PCG algorithm, we generated 1000 random challenges.
For each generated challenge in our trials, we recorded if the challenge was solvable, and if so, we also recorded the minimum solution length found with our A* solver. To enable testing for statistical significance on solvability between these algorithms, we divided our pool of generated challenges into 10 trials of 100 challenges. 
 
\subsubsection{Solvability}

Figure \ref{fig:strategycomp} shows the distribution of solvability for the four PCG algorithms across the 10 trials in a box-and-whisker plot. Each method produces a healthy probability of solvable challenges. {\bf Shuffled} has the lowest mean value for solvability at 85\%, and this result is significantly lower than the other three algorithms, which is confirmed by t-tests using a p-value of 0.05. Within the top three algorithms, only {\bf Dominoes} is statistically higher than {\bf Piecepack}. 

\subsubsection{Interest}

Applying the interest metric of Brown and Maire \cite{MCPUZZLE} across all of our PCG algorithms, we found that $P(S_r)$, the probability that a random solver would win, was less than 0.003, making $P(S_p) - P(S_r)$ equivalent to solvability. As our goal is to determine if the above algorithms change the interest distribution of generated challenges, we consider two other measures found by 
Jaru{\v{s}}ek and Pel{\'a}nek to be correlated with challenge difficulty for one-way transport puzzles:
the minimum number of moves required to solve the challenge, and the number of counter-intuitive moves along the minimum solution path. \cite{jaruvsek2010difficulty}
\cite{jaruvsek2011determines}

First, we determined that a move in Fujisan is counter-intuitive if it involves moving a Priest further away from the summit. For each solvable challenge, we counted the number of counter-intuitive moves used on the minimum solution path length. Using this metric, values ranged from 0-8, and we found that there was no statistical difference between the four algorithms.

Second, we calculated the distribution of minimum solution path length generated by each algorithm. We compared here the median length generated by each algorithm. The shortest possible solution to a Fujisan challenge involves eight moves, while 
the longest-known constructed challenge requires 62 moves\footnote{http://www.ludism.org/ppwiki/Fuji-san\#Heading9}.

\begin{figure}[t]
\centering
\includegraphics[width=8cm]{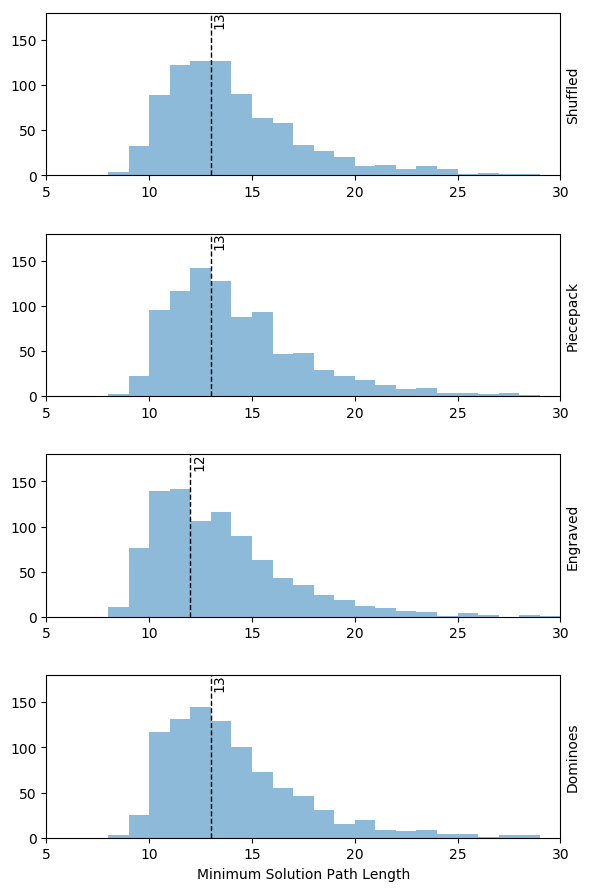}
\caption{Histograms showing the effect of Fujisan challenge generation algorithms on minimum solution length.}
\label{fig:difficultycomp}
\end{figure}

Figure \ref{fig:difficultycomp} shows histograms of the minimum solution length for solvable challenges, pooled across all trials for each algorithm. The median is denoted with a dotted line. Our algorithms appear to follow a Poisson distribution rather than a normal distribution, since the smallest possible solution length for any challenge is 8, and the maximum solution length is currently unbounded. We employ a Kruskal-Wallis H-test \cite{KRUSKAL} to determine if the median length of our four algorithms is statistically the same, and we rejected this null hypothesis very strongly, with a p-value of $6.4 \times 10^{-9}$.

The algorithm responsible for this result was {\bf Engraved Tiles}. We can see a strong tendency to have shorter solution lengths, with almost 10\% of challenges having a solution length of eight or nine, whereas for {\bf Dominoes}, this is true for only 3\% of challenges. In {\bf Engraved Tiles}, there are five tiles that contain a zero value; since there will be ten total tiles hidden, on average a challenge will contain 2.5 zero values. It appears that zero values are one part of what makes Fujisan challenges interesting.

\section{Entangled Components}
\label{entangled}

\noindent
Recall that we define components to become {\it entangled} if their arrangement is no longer independent, either due to algorithmic constraints, or having been subsumed by new, larger components. In each of the explored puzzles above, we see that entangled components can be used to produce challenges with higher solvability without sacrificing the interesting qualities of the puzzle. Here we provide evidence that this is due to aligning the entanglements with desirable portions of the underlying puzzle mechanics. For each puzzle, metrics can be found that are correlated with higher solvability. Entangled components can then be biased toward solvability by encouraging these metrics in generated challenges.

\begin{figure}[t]
\includegraphics[width=8.8cm]{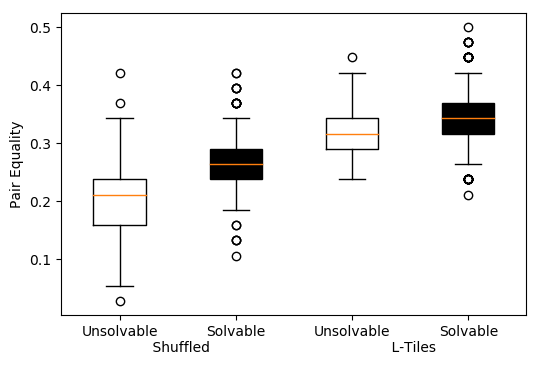}
\caption{Average pair equality for (4, 6, 4) color BoxOff challenges across each setup algorithm. Solvable challenges shown in black, and unsolvable challenges shown in white.}
\label{fig:boxoffconnected}
\end{figure}

In BoxOff, Rule 1 underscores the importance of adjacent grid squares of the same color, as they are the first moves available in the game. The (4, 6, 4) board contains 38 unique pairs of adjacent grid squares. If we randomly distribute the four colors using the \textbf{Shuffled} method, on average a pair will be the same color 22\% of the time. We label this metric {\it pair equality}. However, if we separate the generated challenges by solvability, we see in Figure \ref{fig:boxoffconnected} that while unsolvable challenges average 20\% pair equality, for solvable challenges the average increases to 26\%, and this difference is statistically significant.

Through entanglement, the \textbf{L-Tiles} constrain adjacent pairs such that at least eight of them are the same color, resulting in a minimum of 21\% pair equality. Thus, these challenges are more likely to be in the solvable range, as shown in Figure \ref{fig:boxoffconnected}, with 32\% pair equality for unsolvable challenges, and 34\% pair equality for those that are solvable. To verify that the entangled components must align with the puzzle mechanics, we also tested a configuration of \textbf{L-Tiles} for the (4, 6, 4) puzzle with three unique colors on each tile, evenly distributed among the colors. This resulted in an overall solvability of 8.2\%, much worse than using the original \textbf{Shuffled} algorithm.

To better understand how entangled components improve solvability for Pretzel solitaire, it is revealing to compare the dip in solvability for (4, 3) \textbf{Sequential Suits} with the dramatic spike in solvability for (4, 4) \textbf{Sequential Suits}, both clearly evident in Figure \ref{fig:pretzelfullsolve}. As discussed by de Bruijn, sequences of necessarily dependent moves that turn back on themselves to form loops, which de Bruijn calls cyclic blockades, are the bane of the Pretzel solver. Once a cyclic blockade forms, the Pretzel is no longer solvable. These can either be present immediately following the deal or can develop during the course of play. (4, 3) Pretzels with \textbf{Sequential Suits} are highly prone to at least two shapes of dealt cyclic blockades that are readily detected.

To recognize the first, we observe that exactly three spades will be dealt into the grid in a diagonal line, from the first column of the first row to the third column of the third row. There is a 1 in 6 chance that the $3\spadesuit$ will be dealt into the first column of the first row and the $2\spadesuit$ into the third column of the third row. When this occurs, the $2\spadesuit$ can never move into its goal location, occupied by the $3\spadesuit$, which itself cannot move into a hole following the $2\spadesuit$, since the $2\spadesuit$ is in the rightmost column and is unable to leave that column. We shall call this shape of cyclic blockade a Ducking Crab.

So, at least 1 in 6 (4, 3) Pretzels with \textbf{Sequential Suits} are unsolvable immediately after setup. Notice, however, that a Ducking Crab is just as probable for diamonds, so at least 1 in 3 are immediately unsolvable.

A second shape of cyclic blockade that plagues \textbf{Sequential Suits} for the (4, 3) case occurs when the $2\clubsuit$ and $2\heartsuit$ are dealt into each other's goal locations, preventing either from ever moving. Such Duelling Deuces occur in 1 of 9 challenges.

Ducking Crabs and Duelling Deuces are not, of course, exclusive to (4, 3) Pretzels with \textbf{Sequential Suits}. Nor have we attempted to identify and enumerate an exhaustive list of every possible shape for cyclic blockades. We offer Ducking Crabs and Duelling Deuces merely as examples of the kinds of structures that limit solvability for Pretzels. They are suitable examples not only because they account for a significant portion of unsolvable (4, 3) Pretzels with \textbf{Sequential Suits}, but also because it is plain to see how both are completely avoided by \textbf{Sequential Suits} for the (4, 4) case and indeed for all cases $S = V$, as well as \textbf{Banded Suits}. With a stripe of spades filling the first column, there can only ever be a single 2 in the column, preventing Deulling Deuces. Although the $3\spadesuit$ might be dealt into $2\spadesuit$'s goal location, the $2\spadesuit$ can never be dealt into the rightmost column, so Ducking Crabs are likewise impossible. Thus for Pretzel solitaire, entangled components help to avoid some, but not all, of the structures that destroy solvability.

For Fujisan, the entangled components are a direct result of the puzzle movement mechanics. We first develop a metric called {\it connectivity} based on Rule 2 to explore the connections between steps in a challenge. We say step $A$ is connected to step $B$ in Fujisan if there is a move available according to Rule 2 from $B$ to $A$. This metric, only an approximation of true puzzle connectivity as it ignores connections possible from Rule 2.a with intermediate Priests, contains enough information to bias our entangled components.

\begin{figure}[t]
\includegraphics[width=8.8cm]{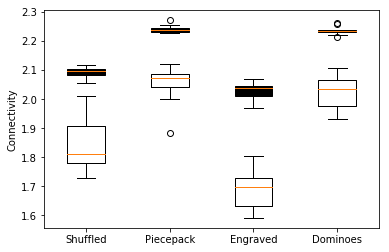}
\caption{Average connectedness for Fujisan challenges across each setup algorithm. 
Solvable challenges shown in black, and unsolvable challenges shown in white.}
\label{fig:connected}
\end{figure}

Figure \ref{fig:connected} shows the average step connectivity within a challenge for each setup algorithm, differentiating solvable challenges in black from unsolvable challenges in white. 
In the \textbf{Shuffled} algorithm, we can see a large divide between solvable and unsolvable challenges, and across all algorithms, higher connectivity is always related to higher solvability. Also, both \textbf{Piecepack} and \textbf{Dominoes} require that each step has two unique values. In these two algorithms, this uniqueness constraint strongly increases the connectivity of both solvable and unsolvable challenges, but the divide remains intact.

However, even if each step is well-connected to other steps, the connections could form patterns and loops between the steps, thus breaking the potential Priest movement into disjoint sets. If we add a restriction that each step pair in a challenge be a unique set of values, this will cause the connections between steps to be more distributed and bind the puzzle together as a whole. In our experiments, we found that Shuffled challenges which fit this restriction can be solved at a rate of 89\%, a statistical improvement over the general Shuffled challenge population. As we noted earlier, the {\bf Engraved Tiles} and {\bf Dominoes} incorporate this prohibition on the repetition of steps as they subsume the coins into larger components. In fact, the most successful algorithm, {\bf Dominoes}, combines both of these constraints to create well-connected and well-distributed challenges.

\section{Conclusion}   \label{sec:Conclusion}

\noindent
Our work introduces the idea of entangled components for physical solitaire puzzle games, and explores their implications for procedural content generation algorithms across three puzzles. We can see that subtle changes in the game components can affect their random distribution, leading to large-scale changes on the generated challenges. When aligned with the puzzle mechanics, entangled components lead to increased challenge solvability without sacrificing the interesting nature of the puzzle.

There are many open questions related to physical games and PCG. First, we believe there is work to be done in formalizing, and validating with human subjects, an ease of physical setup metric. While the \textbf{Shuffled} algorithm for each puzzle is very simple to execute, some of the more entangled algorithms could be time-consuming and tedious for a human to implement. A simple approximation metric would be the time complexity of the algorithm; however, certain operations that are straightforward to a computer can be difficult for humans to track, and vice versa. With a formal metric, game designers could be inclined to include more intricate PCG algorithms when provided guarantees these algorithms can reasonably be executed by a human player.

A further point to clarify is the exact relationship between challenge interest and challenge difficulty. We focused here on how entangled components maintained challenge interest, but this simplifies the potential for gradations of challenge difficulty within each puzzle. Validating a sophisticated difficulty metric for each puzzle, again with human subject tests as disussed in Jaru{\v{s}}ek and Pel{\'a}nek \cite{jaruvsek2011determines}
, could further illustrate the effects of entangled components.

Finally, are there general entangled methods that allow solvers to {\it construct} challenges online to guarantee solvability, as opposed to the {\it generate-and-test} algorithms discussed here? While this may be possible in certain situations, 
care must be taken that the construction process does not give away the solution to the challenge.

\ifCLASSOPTIONcaptionsoff
  \newpage
\fi

\bibliographystyle{IEEEtran}
\bibliography{references.bib}

\begin{IEEEbiography}[{\includegraphics[width=1in,height=1.25in,clip,keepaspectratio]{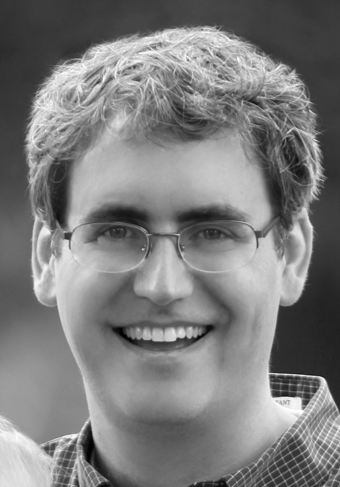}}]{Mark Goadrich} earned an A.B. in Mathematics and Philosophy at Kenyon College, and a M.S. and Ph.D. in Computer Sciences from the University of Wisconsin - Madison.
He is currently an Associate Professor of Computer Science at Hendrix College in Conway, AR.
\end{IEEEbiography}

\begin{IEEEbiography}[{\includegraphics[width=1in,height=1.25in,clip,keepaspectratio]{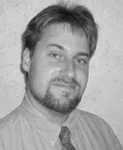}}]{James Droscha}
is a designer and developer of software, games, and puzzles, including the piecepack and Fujisan.
\end{IEEEbiography}

\vfill

\end{document}